\documentclass[final]{cvpr}

\usepackage{times}
\usepackage{graphicx}
\usepackage{comment}
\usepackage{amsmath,amssymb} 
\usepackage{color}

\usepackage[dvipsnames, table]{xcolor}
\usepackage{enumitem}

\usepackage[square,numbers,sort&compress]{natbib}
\usepackage{algpseudocode}
\usepackage[font=small,labelfont=bf]{caption}
\usepackage{array}
\usepackage{multirow}
\usepackage{booktabs}
\usepackage[linesnumbered,ruled,vlined]{algorithm2e}
\usepackage{subcaption}
\usepackage[normalem]{ulem}
\usepackage{xparse}
\usepackage{pifont}
\usepackage{bm}
\usepackage{epigraph}

\usepackage{tabularx}
\usepackage{ifthen}
\usepackage{mwe} 
\usepackage{makecell}
\usepackage{color, colortbl}

\usepackage{wrapfig}

\usepackage[scaled=0.85]{DejaVuSansMono}

\usepackage[pagebackref=true,breaklinks=true,colorlinks,citecolor=ForestGreen,bookmarks=false]{hyperref}

\usepackage{cleveref}
\crefformat{section}{\S#2#1#3}
\crefformat{subsection}{\S#2#1#3}
\crefformat{subsubsection}{\S#2#1#3}
\crefformat{figure}{Figure~#2#1#3}
\crefformat{equation}{Equation~#2#1#3}
\crefformat{table}{Table~#2#1#3}
\crefformat{algorithm}{Algorithm~#2#1#3}

\SetCommentSty{mycommfont}
\SetKwInput{KwInput}{Input}                

\usepackage{microtype}

\setlist{nosep} 

\newcommand{\NCE}{NCE\xspace}

\newcommand{\AVID}{AVID\xspace}
\newcommand{\AVIDName}{Audio-Visual Instance Discrimination\xspace}

\newcommand{\CMA}{CMA\xspace}
\newcommand{\CMAName}{Cross-Modal Agreement\xspace}

\newcommand{\x}{\mathbf{x}}

\renewcommand{\a}{\mathbf{a}}

\renewcommand{\v}{\mathbf{v}}

\renewcommand{\L}{\mathcal{L}}
\newcommand{\Lnce}{\mathcal{L}_{\mbox{\footnotesize NCE}}}

\newlength\savewidth
\newlength\thinwidth

\definecolor{Gray}{gray}{0.93}
\newcolumntype{a}{>{\columncolor{Gray}}c}
\definecolor{LightCyan}{rgb}{0.88,1,1}
\definecolor{highlightRowColor}{gray}{0.93}
\definecolor{HighlightBlue}{RGB}{230, 235, 247}

\newcommand{\mc}[3]{\multicolumn{#1}{#2}{#3}}
\newcommand{\mr}[2]{\multirow{#1}{*}{#2}}

\newcommand{\HC}[1]{\ifthenelse{\isodd{#1}}{\rowcolor{highlightRowColor}}{\rowcolor{white}}}
\newcommand{\rcw}{\rowcolor{white}}
\newcommand{\rcg}{\rowcolor{highlightRowColor}}

\usepackage{xspace} 
\makeatletter
\DeclareRobustCommand\onedot{\futurelet\@let@token\@onedot}
\def\@onedot{\ifx\@let@token.\else.\null\fi\xspace}

\def\ie{\emph{i.e}\onedot} 
 
\def\etc{\emph{etc}\onedot} \def\vs{\emph{vs}\onedot}
 
\def\etal{\emph{et al}\onedot}
\makeatother

\usepackage{microtype}

\def\cvprPaperID{1704} 
\def\confYear{CVPR 2021}

\begin{document}

\title{Audio-Visual Instance Discrimination with Cross-Modal Agreement}

\author{Pedro Morgado\thanks{Work done while interning at Facebook AI Research.}\\
UC San Diego
\and
Nuno Vasconcelos\\
UC San Diego
\and
Ishan Misra\\
Facebook AI Research
}

\maketitle
\pagestyle{empty}
\thispagestyle{empty}

\begin{abstract}
We present a self-supervised learning approach to learn audio-visual representations from video and audio. Our method uses contrastive learning for cross-modal discrimination of video from audio and vice-versa. We show that optimizing for cross-modal discrimination, rather than within-modal discrimination, is important to learn good representations from video and audio. With this simple but powerful insight, our method achieves highly competitive performance when finetuned on action recognition tasks. Furthermore, while recent work in contrastive learning defines positive and negative samples as individual instances, we generalize this definition by exploring cross-modal agreement. 
We group together multiple instances as positives by measuring their similarity in both the video and audio feature spaces. Cross-modal agreement creates better positive and negative sets, which allows us to calibrate visual similarities by seeking within-modal discrimination of positive instances, and achieve significant gains on downstream tasks.

\end{abstract}

\section{Introduction}

Imagine the sound of waves. This sound can evoke the memory of many scenes - a beach, a pond, a river, \etc 
A single sound serves as a bridge to connect multiple sceneries. It can group visual scenes that `go together', and set apart the ones that do not. We leverage this property of freely occurring audio to learn video representations in a self-supervised manner. 

A common technique~\cite{owens, owens2016ambient, bruno_avts, l3} is to setup a verification task that requires predicting if an input pair of video and audio is `correct' or not. A correct pair is an `in-sync' video and audio and an incorrect pair can be constructed by using `out-of-sync' audio~\cite{bruno_avts} or audio from a different video~\cite{l3}. However, a task that uses a \emph{single} pair at a time misses a key opportunity to reason about the data distribution at large.

In our work, we propose a contrastive learning framework to learn cross-modal representations in a self-supervised manner by contrasting video representations against \emph{multiple} audios at once (and vice versa). We leverage recent advances~\cite{instance,hadsell2006dimensionality,cmc,oord2018representation} in contrastive learning to setup a \AVIDName (\AVID) task that learns a cross-modal similarity metric by grouping video and audio \emph{instances} that co-occur. We show that the cross-modal discrimination task, \ie, predicting which audio matches a video, is more powerful than the within-modal discrimination task, predicting which video clips are from the same video. With this insight, our technique learns powerful visual representations that improve upon prior self-supervised methods on action recognition benchmarks like UCF-101~\cite{ucf} and HMDB-51~\cite{hmdb}.

We further identify important limitations of the \AVID task and propose improvements that allow us to 1) reason about \emph{multiple} instances and 2) optimize for visual similarity rather than just cross-modal similarity. We use \CMAName (\CMA) to group together videos with high similarity in video and audio spaces. This grouping allows us to directly relate multiple videos as being semantically similar, and thus directly optimize for visual similarity in addition to cross-modal similarity. We show that \CMA can identify semantically related videos, and that optimizing visual similarity among related videos significantly improves the learned visual representations. 
Specifically, \CMA is shown to improve upon AVID on action recognition tasks such Kinetics~\cite{kinetics}, UCF-101~\cite{ucf} and HMDB-51~\cite{hmdb} under both linear probing and full fine-tuning evaluation protocols.

\begin{figure*}[t!]
    \centering
    \includegraphics[width=0.95\textwidth]{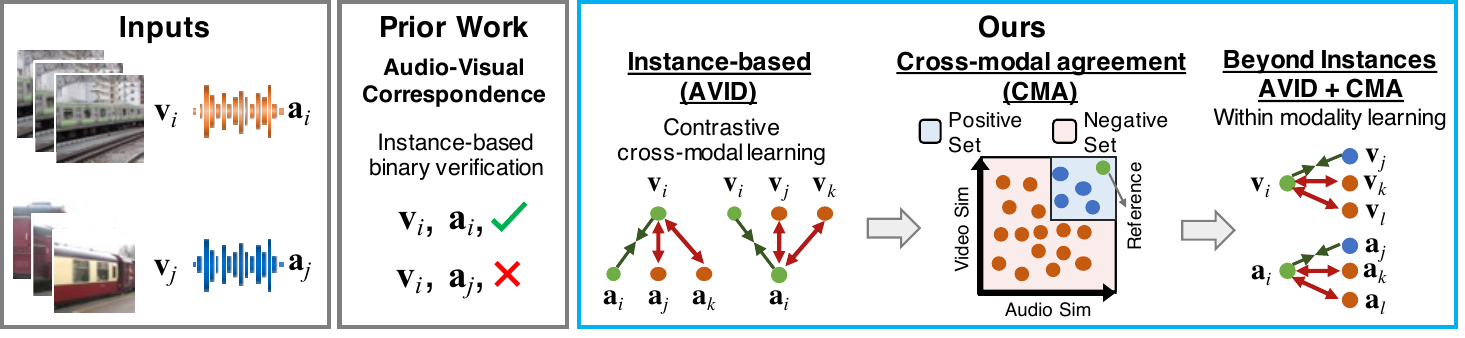}
    \caption{Popular audio-video self-supervised methods can be interpreted as `instance-based' as they learn to align video and audio \emph{instances} by solving a binary verification problem. We propose \AVID to learn cross-modal representations that align video and audio instances in a contrastive learning framework. However, \AVID does not optimize for visual similarity. We calibrate \AVID by formulating \CMA. \CMA finds groups of videos that are similar in both video and audio space which enables us to directly optimize representations for visual (within modality) similarity by using these groups.}
    \label{fig:teaser}
\end{figure*}

\section{Related work}
\label{sec:related_work}

Self-supervised learning is a well studied problem~\cite{olshausen1996emergence,mobahi2009deep,masci2011stacked,salakhutdinov2009deep,de1994learning,le2013building}. Self-supervised methods often try to reconstruct the input data or impose constraints on the representation, such as sparsity~\cite{lee2007efficient,olshausen1996emergence,olshausen2000sparse}, noise~\cite{vincent2008extracting} or invariance~\cite{hadsell2006dimensionality,ranzato2007unsupervised,dosovitskiy2016discriminative,ji2018invariant,misra2019pirl,bojanowski2017unsupervised,deepcluster,caron2019deep} to learn a useful and transferable feature representation. 
An emerging area of research uses the structural or domain-specific properties of visual data to algorithmically define `pretext tasks'. Pretext tasks are generally not useful by themselves and are used as a proxy to learn semantic representations. They can use the spatial structure in images~\cite{doersch2015unsupervised,jigsaw,rotation,zhang2019aet}, color~\cite{colorization,larsson2016learning,larsson2017colorization,deshpande2015learning}, temporal information in videos~\cite{shuffle,opn,fernando2017self,3d_rotation,pathak2017learning,counting,wang2015unsupervised,dpc,dwibedi2019tcc} among other sources of `self' or naturally available supervision. We propose an unsupervised learning technique that leverages the naturally available signal in video and audio alignment.

\paragraph{Representation Learning using Audio.} Self-supervised learning can also make use of multiple modalities, rather than the visual data alone. As pointed out in~\cite{de1994learning,kidron2005pixels}, co-occurring modalities such as audio can help learn powerful representations. For example, audio self-supervision has shown to be useful for sound source localization and separation~\cite{arandjelovic2018objects,avcloc_senocak2018,gao2018learning,gao2019co,zhao2018sound,zhao2019sound,gan2020music}, lip-speech synchronization~\cite{chung2016out} and visual representation learning~\cite{l3,bruno_avts,owens} and audio spatialization~\cite{morgado2018self}.

\paragraph{Audio-Visual Correspondence (AVC)} is a standard task~\cite{l3, bruno_avts,owens,arandjelovic2018objects} used in audio-video cross-modal learning. This task tries to align the visual and audio inputs by solving a binary classification problem. However, most methods use only a single video and single audio at a time for learning. Thus, the model must reason about the distribution over multiple samples implicitly. In our work, we use a contrastive loss~\cite{hadsell2006dimensionality,oord2018representation,cmc,instance} that opposes a large number of samples simultaneously. We show in~\cref{sec:experiments} that our method performs better than recent methods that use AVC.

\paragraph{Contrastive Learning} techniques use a contrastive loss~\cite{hadsell2006dimensionality} to learn representations either by predicting parts of the data~\cite{oord2018representation,henaff2019data,hjelm2018learning}, or discriminating between individual training instances~\cite{instance,dosovitskiy2016discriminative,moco,misra2019pirl,local_aggregation,feng2019self,ye2019unsupervised,AdvAugmCL}. Contrastive learning has also been used for learning representations from video alone~\cite{dpc,sermanet2018time}. 
Tian \etal~\cite{cmc} also use a contrastive approach, but propose to learn with a cross-modal objective applied to images and depth, video and flow. 
In contrast, our method learns visual representations using audio as cross-modal targets.
Compared to~\cite{cmc}, we present a new insight for audio-visual learning that optimizing cross-modal similarity is more beneficial than within-modal similarity. We also identify important limitations of cross-modal discrimination and present an approach that goes beyond instance discrimination by modeling \CMAName. This identifies groups of related videos and allows us to optimize for within-modal similarity between related videos. The concurrently proposed~\cite{xmodalclust} uses alternating optimization to find clusters in visual and audio feature spaces, \textit{independently} and uses them to improve cross-modal features. While our \CMA method bears a resemblance to theirs, we do not use alternating optimization and use \textit{agreements} between the visual and audio representations to directly improve visual similarity rather than only cross-modal similarity.
Finally, similar to our work, the concurrently proposed~\cite{CoCLR} also uses co-occurring modalities (optical flow and RGB) to expand the positive set. However, instead of mining positives based on an agreement between both modalities, \cite{CoCLR} relies on the opposite modality alone.

\paragraph{Multi-view Learning.} Multi-view learning aims to find common representations from multiple views of the same phenomenon, and has been widely used to provide learning signals in unsupervised and semi-supervised applications. Classical approaches can be broadly categorized in co-training procedures~\cite{cotraining, bickel_mvcluster, wang2007analyzing, kumar_mv_spectclust, ma2017self, qiao2018deep, CoCLR} that maximize the mutual agreement between views, multiple kernel learning procedures~\cite{lanckriet2004learning, bach2004multiple, kloft2011local} which use kernels to model different views, and subspace learning procedures~\cite{diethe2008multiview, quadrianto2011learning} which seek to find the latent space that generates all views of the data.

Multi-view data is an effective source of supervision for self-supervised representation learning. Examples include the motion and appearance of a video~\cite{cmc,CoCLR}, depth and appearance~\cite{splitbrain, jiang2018self}, luminance and chrominance of an image~\cite{splitbrain, cmc}, or as in our work sound and video~\cite{l3,soundnet,owens2016ambient,chung2016out}. 
\section{Audio-Visual Instance Discrimination}
\label{sec:AVID}

\begin{figure*}[t!]
    \centering
    \includegraphics[width=0.7\linewidth]{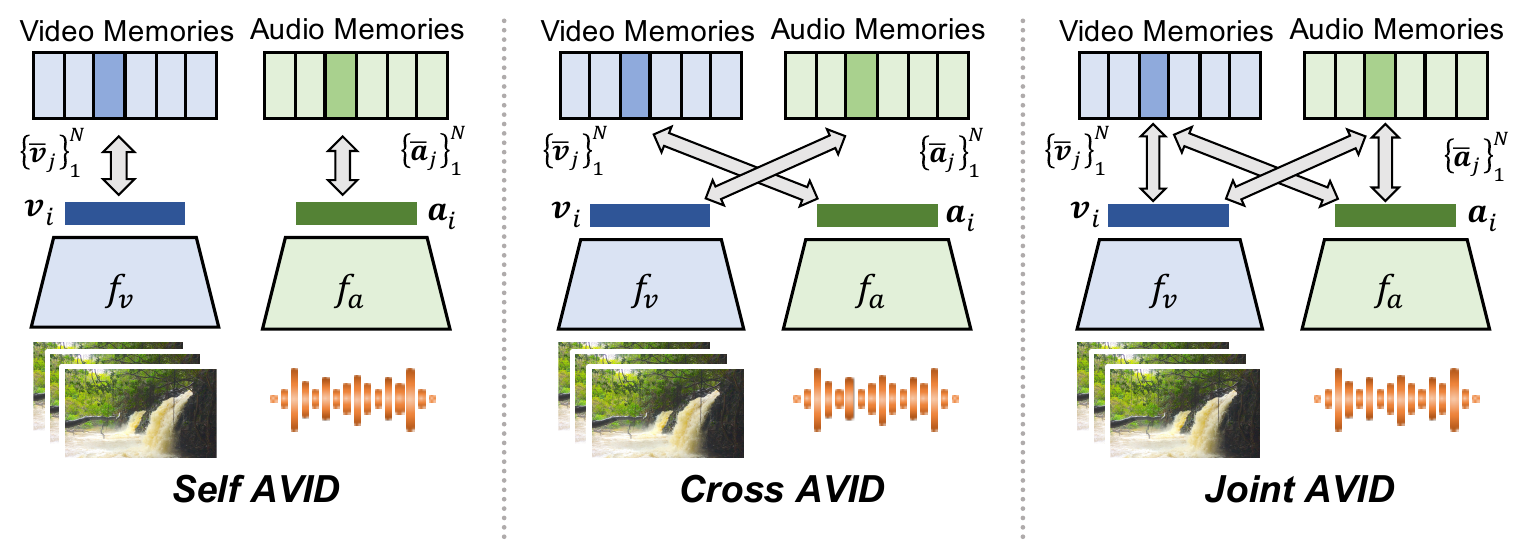}
    \caption{\textbf{Variants of the \AVID task.} Instance discrimination can be accomplished contrasting representations within the same modality (Self-\AVID), across modalities (Cross-\AVID) or a mixture of the two (Joint-\AVID).
    }
    \label{fig:AVID}
\end{figure*}

We learn visual representations in a self-supervised manner from unconstrained video and audio by building upon recent advances in instance discrimination~\cite{instance,dosovitskiy2016discriminative,malisiewicz2011ensemble,cmc} and contrastive learning~\cite{hadsell2006dimensionality,nce,oord2018representation}. 

\subsection{Goal and Intuition.} 
Consider a dataset of $N$ samples (instances) $\mathcal{S}=\{s_i\}_{i=1}^{N}$ where each instance $s_i$ is a video $s^v_i$ with a corresponding audio $s^a_i$. The goal of \AVIDName (\AVID) is to learn visual and audio representations $(\v_i, \a_i)$ from the training instances $s_i$. The learned representations are optimized for `instance discrimination'~\cite{dosovitskiy2016discriminative,instance,malisiewicz2011ensemble}, \ie, must be discriminative of $s_i$ itself as opposed to other instances $s_j$ in the training set. Prior work~\cite{dosovitskiy2016discriminative,instance} shows that such a discriminative objective among instances learns semantic representations that capture similarities between the instances.

To accomplish this, two neural networks extract unit norm feature vectors $\v_i=f_v(s_i^{v})$ and $\a_i=f_a(s_i^{a})$ from the video and audio independently. Slow moving (exponential moving average) representations for both video and audio features $\{(\bar\v_i, \bar\a_i)\}_{i=1}^N$ are maintained as `memory features' and used as targets for contrastive learning. The \AVID task learns representations $(\v_i, \a_i)$ that are more similar to the memory features of the instance $(\bar\v_i, \bar\a_i)$ as opposed to memory features of other instances $(\bar\v_j, \bar\a_j)$, $j\!\neq\!i$.
However, unlike previous approaches~\cite{instance,dosovitskiy2016discriminative} defined on a single modality (but similar to~\cite{cmc}), \AVID uses multiple modalities, and thus can assume multiple forms as depicted in~\cref{fig:AVID}.
\begin{enumerate}[leftmargin=*]
    \item \textbf{Self-\AVID} requires instance discrimination within the same modality - $\v_i$ to $\bar\v_i$ and $\a_i$ to $\bar\a_i$. This is equivalent to prior work~\cite{dosovitskiy2016discriminative,instance} independently applied to the two modalities.
    \item \textbf{Cross-\AVID} optimizes for cross-modal discrimination, \ie, the visual representation $\v_i$ is required to discriminate the accompanying audio memory $\bar\a_i$ and vice-versa.
    \item \textbf{Joint-\AVID} combines the Self-\AVID and Cross-\AVID objectives.
\end{enumerate}
It is not immediately obvious what the relative advantages, if any, of these variants are. In \cref{sec:analysis-avid}, we provide an in-depth empirical study of the impact of these choices on the quality of the learned representations. We now describe the training procedure in detail.

\subsection{\AVID training procedure.}
\AVID is trained using a contrastive learning framework~\cite{hadsell2006dimensionality,nce}, where instance representations are contrasted to those of other (negative) samples.

While various loss functions have been defined  for contrastive learning~\cite{oord2018representation,schroff2015facenet}, we focus on noise contrastive estimation (NCE)~\cite{nce}. Let $\bar\x_i$ denote the (memory) target representation for a sample $s_i$.
The probability that a feature $\x$ belongs to sample $s_i$ is modeled by a generalized softmax function
\begin{equation}
    P(s_i |\x) = \tfrac{1}{N\bar{Z}}\exp(\x^T \bar\x_i / \tau)
    \label{eq:softmax}
\end{equation}
where $\bar{Z} = \tfrac{1}{N}\sum_{\bar\x}[\exp(\x^T \bar\x / \tau)]$ 
is the normalized partition function and $\tau$ is a temperature hyper-parameter that controls the softness of the distribution. In the case of \AVID, $\x$ and $\bar\x$ may or may not be from the same modality.

\newcommand{\VideoKineticsAVIDTable}{
\begin{tabular}{c|cccc|c}
\toprule
Method & \bf block1 & \bf block2 & \bf block3 & \bf block4 & \bf Best \\
\hline
\rcg Cross-AVID & \bf 19.80 & \bf 26.98 & \bf 34.81 & \bf 39.95 & \bf 39.95 \\
\rcw Self-AVID  & 17.10 & 22.28 & 27.23 & 32.08 & 32.08 \\
\rcg Joint-AVID & 18.65 & 23.60 & 29.47 & 33.04 & 33.04 \\
\bottomrule
\end{tabular}
}
\newcommand{\ESCVideoAVIDTable}{
\begin{tabular}{c|cccc|c}
\toprule
& \bf block1 & \bf block2 & \bf block3 & \bf block4 & \bf Best \\
\hline
\rcg Cross-AVID & \bf 67.25 & \bf 73.15 & \bf 74.80 & \bf 75.05 & \bf 75.05 \\
\rcw Self-AVID & 66.92 & 72.64 & 71.45 & 71.61 & 72.64 \\
\rcg Joint-AVID & 65.45 & 68.65 & 71.77 & 68.41 & 71.77 \\
\bottomrule
\end{tabular}
}
\begin{table*}[t!]
    \centering
    \captionsetup[subtable]{position=below}
    \captionsetup[table]{position=top}
    \caption{{\bf Variants of \AVID.} We observe that the Self-\AVID and Joint-\AVID variants that use within-modality instance discrimination perform poorly compared to Cross-\AVID that uses only cross-modal instance discrimination.
    }
    \label{tab:analysis-avid}
    \begin{subtable}{0.43\linewidth}
        \centering
        \resizebox{\linewidth}{!}{\VideoKineticsAVIDTable}
        \caption{Accuracy of linear probing on Kinetics.}
    \end{subtable}%
    \qquad
    \begin{subtable}{0.43\linewidth}
        \centering
        \resizebox{\linewidth}{!}{\ESCVideoAVIDTable}
        \caption{Accuracy of linear probing on ESC.}
    \end{subtable}
    \vspace{-2pt}
\end{table*}

The network $f$ is trained to learn representations by solving multiple binary classification problems where it must choose its own target representation $\bar\x_i$ over representations $\bar\x_j$ in a negative set. The negative set consists of $K$ `other' instances drawn uniformly from $\mathcal{S}$, \ie, 
$\mathcal{N}_{i}=\mathcal{U}(\mathcal{S})^K$.
The probability of a feature $\x$ being from instance $s_i$ as opposed to the instances from the uniformly sampled negative set $\mathcal{N}_{i}$ is given as
$
    P(D=1 | \x, \bar\x_i) = 
    \frac{P(s_i|\x)}{P(s_i|\x) + K/N}. 
$
The \NCE loss is defined as the negative log-likelihood
\begin{align}
    \Lnce(\x_i; \bar\x_i, \mathcal{N}_{i})=
    & - \log P(D=1 | \x_i, \bar\x_i) \nonumber \\
    & - \sum_{j\in\mathcal{N}_{i}}\log P(D=0 | \x_i, \bar\x_j),
    \label{eq:nce_loss}
\end{align}
where $P(D=0 |\cdot) = 1-P(D=1 |\cdot)$.

The three variants of \AVID depicted in \cref{fig:AVID} are trained to optimize variations of the NCE loss of~\cref{eq:nce_loss}, by varying the target representations~$\bar\x_i$.
\begin{align}
    \L_{\mbox{\tiny Self-AVID}} = \L_{\mbox{\tiny NCE}}(\v_i; \bar\v_i, \mathcal{N}_i) + \L_{\mbox{\tiny NCE}}(\a_i; \bar\a_i, \mathcal{N}_i)
    \label{eq:self_avid_loss} \\
    \L_{\mbox{\tiny Cross-AVID}} = \L_{\mbox{\tiny NCE}}(\v_i; \bar\a_i, \mathcal{N}_i) + \L_{\mbox{\tiny NCE}}(\a_i; \bar\v_i, \mathcal{N}_i)
    \label{eq:cross_avid_loss} \\
    \L_{\mbox{\tiny Joint-AVID}} = \L_{\mbox{\tiny Self-AVID}}(\v_i, \a_i) + \L_{\mbox{\tiny Cross-AVID}}(\v_i, \a_i)
    \label{eq:joint_avid_loss}
\end{align}

We analyze these variants next and show that the seemingly minor differences between them translate to significant differences in performance.


\subsection{Analyzing \AVID}
\label{sec:analysis-avid}

We present experiments to analyze various properties of the \AVID task and understand the key factors that enable the different variants of \AVID to learn good representations.

\par \noindent \textbf{Experimental Setup}
We briefly describe the experimental setup for analysis and provide the full details in the supplemental.

\par \noindent \emph{Pre-training Dataset.}
All models are trained using the Audioset dataset~\cite{audioset} which contains 1.8M videos focusing on audio events. We randomly subsample 100K videos from this dataset to train our models. We use input video and audio clips of 1 and 2-second duration, respectively. The video model is trained on 16 frames of size $112\!\times\!112$ with standard data augmentation~\cite{szegedy2015going}. We preprocess the audio by randomly sampling the audio within 0.5 seconds of the video and compute a log spectrogram of size $100\!\times\!129$ (100 time steps with 129 frequency bands).

\par \noindent \emph{Video and audio models.}
The video model is a smaller version of the R(2+1)D models proposed in~\cite{tran2018closer} with 9 layers. The audio network is a 9 layer 2D ConvNet with batch normalization.
In both cases, output activations are max-pooled, projected into a 128-dimensional feature using a multi-layer perceptron (MLP), and normalized into the unit sphere. The MLP is composed of three fully connected layers with 512 hidden units.

\par \noindent \emph{Pre-training details.}
\AVID variants are trained to optimize the loss in Equations~\ref{eq:self_avid_loss}-\ref{eq:joint_avid_loss} with 1024 random negatives. In early experiments, we increased the number of negatives up to 8192 without seeing noticeable differences in performance.
Following~\cite{instance}, we set the temperature hyper-parameter $\tau$ to 0.07, the EMA update constant to 0.5, and the normalized partition function $\bar Z$ is approximated during the first iteration and kept constant thereafter ($\bar Z=2.2045$).
All models are trained with the Adam optimizer~\cite{adam} for 400 epochs with a learning rate of 1e-4, weight decay of 1e-5, and batch size of 256.

\par \noindent \emph{Downstream tasks.}
We evaluate both the visual and audio features using transfer learning.

\begin{itemize}[leftmargin=*]
    \setlength\itemsep{0pt}
    \item Visual Features: We use the Kinetics dataset~\cite{kinetics} for action recognition. We evaluate the pre-trained features by linear probing~\cite{scaling,splitbrain} where we keep the pre-trained network fixed and train linear classifiers. 
    We report top-1 accuracy on held-out data by averaging predictions over 25 clips per video.
    \item Audio Features: We evaluate the audio features on the ESC-50~\cite{esc} dataset by training linear classifiers on fixed features from the pre-trained audio network. Similar to the video case, we report top-1 accuracy by averaging predictions over 25 clips per video.
\end{itemize}

\par \noindent \textbf{Cross \vs within-modal instance discrimination}

We study the three variants of \AVID depicted in~\cref{fig:AVID} to understand the differences between cross-modal and within-modal instance discrimination and its impact on the learned representations. 
We evaluate the video and audio feature representations from these variants and report results in~\cref{tab:analysis-avid}. We observe that Self-\AVID is consistently outperformed by the Cross-\AVID variant on both visual and audio tasks. 

We believe the reason is that Self-\AVID uses within-modality instance discrimination, which is an easier pretext task and can be partially solved by matching low-level statistics of the data~\cite{doersch2015unsupervised,l3}. This hypothesis is supported by the fact that Joint-\AVID, which combines the objectives of both Cross-\AVID and Self-\AVID, also gives worse performance than Cross-\AVID.
These results highlight that one \emph{cannot} naively use within-modality instance discrimination when learning audio-visual representations.
In contrast, Cross-\AVID uses a ``harder'' cross-modal instance discrimination task where the video features are required to match the corresponding audio and vice-versa. As a result, it generalizes better to downstream tasks.

\section{Beyond Instance Discrimination: \CMAName}

We will show in~\cref{sec:experiments} that Cross-\AVID achieves state-of-the-art performance on action recognition downstream tasks. However, we identify three important limitations in the instance discrimination framework of~\cref{eq:nce_loss} and the cross-modal loss of \cref{eq:cross_avid_loss}.
\begin{enumerate}[leftmargin=*]
    \setlength\itemsep{0pt}
    \item \textbf{Limited to instances:} Instance discrimination does not account for interactions \emph{between} instances. Thus, two semantically related instances are never grouped together and considered `positives'.
    \item \textbf{False negative sampling:} The negative set $\mathcal{N}_i$, which consists of \emph{all} other instances $s_j$, may include instances semantically related to $s_i$. To make matters worse, contrastive learning requires a large number $K$ of negatives, increasing the likelihood that semantically related samples are used as negatives. This contradicts the goal of representation learning, which is to generate similar embeddings of semantically related inputs.
    \item \textbf{No within-modality calibration:} The Cross-\AVID loss of Equation~\ref{eq:cross_avid_loss} does not directly optimize for visual similarity $\v_i^T\v_j$. In fact, as shown experimentally in~\cref{sec:analysis-avid}, doing so can significantly hurt performance. Nevertheless, the lack of within-modality calibration is problematic, as good visual representations should reflect visual feature similarities.
\end{enumerate}

\subsection{Relating instances through agreements}
\label{sec:CMA}

We extend \AVID with \CMAName (\CMA) to address these shortcomings. \CMA builds upon insights from prior work~\cite{rosenberg2005semi} in multi-view learning. We hypothesize that, if two samples are similar in \emph{both} visual and audio feature space, then they are more likely to be semantically related than samples that agree in only one feature space (or do not agree at all). We thus consider instances that agree in both feature spaces to be `positive' samples for learning representations. Similarly, examples with a poor agreement in either (or both) spaces are used as negatives. When compared to instance discrimination methods~\cite{instance,cmc,dosovitskiy2016discriminative}, \CMA uses a larger positive set of semantically related instances and a more reliable negative set. 


\subsection{\CMA Learning Objective}

We define an agreement score for two instances $s_i$ and $s_j$ as
\begin{equation}
    \rho_{ij} = \min(\v_i^T\v_j, \a_i^T\a_j).
\end{equation}
This is large only when {\it both\/} the audio and video similarities are large. A set of positives and negatives is then defined per instance $s_i$. 
The positive set $\mathcal{P}_i$ contains the samples that are most similar to $s_i$ in both spaces, while the negative set $\mathcal{N}_i$ is the complement of $\mathcal{P}_i$.
\begin{equation}
    \mathcal{P}_i = \underset{j=1,\ldots,N}{\mbox{TopK}}(\rho_{ij}) 
    \qquad
    \mathcal{N}_i = \{j | s_j\in (\mathcal{S} \setminus \mathcal{P}_i) \} \label{eqn:cma_sets}
\end{equation}

Furthermore, \CMA enables self-supervision beyond single instances.
This is achieved with a generalization of the \AVID task, which accounts for the correspondences of \cref{eqn:cma_sets}.  At training time, $K_n$ negative instances are drawn per sample $s_i$ from the associated negative set $\mathcal{N}_i$ to form set $\mathcal{N}^\prime_i=\mathcal{U}(\mathcal{N}_i)^{K_n}$. 
The networks $f_v, f_a$ are learned to optimize a combination of \textit{cross-modal instance discrimination} and \textit{within-modal positive discrimination} (wMPD). 
The former is encouraged through the Cross-\AVID loss of Equation~\ref{eq:cross_avid_loss}.
The latter exploits the fact that \CMA defines \emph{multiple} positive instances $\mathcal{P}_i$, thus enabling the optimization of within-modality positive discrimination
\begin{equation}
    \L_{\mbox{\tiny wMPD}} =
    \frac{1}{K_p} \sum_{p\in\mathcal{P}_i}
    \L_{\mbox{\tiny NCE}}(\v_i; \bar\v_p, \mathcal{N}^\prime_i) + \L_{\mbox{\tiny NCE}}(\a_i; \bar\a_p, \mathcal{N}^\prime_i).
    \label{eq:wm_nce}
\end{equation}
Note that, unlike the Self-\AVID objective of Equation~\ref{eq:self_avid_loss}, this term calibrates within-modal similarities between positive samples. This avoids within-modal comparisons to the instance itself, which was experimentally shown to produce weak representations in~\cref{sec:analysis-avid}.
We then minimize the weighted sum of the two losses
\begin{equation}
    \L_{\mbox{\footnotesize \CMA}} = \L_{\mbox{\footnotesize Cross-\AVID}}(\v_i, \a_i) + \lambda \L_{\mbox{\footnotesize wMPD}}(\v_i, \a_i),
    \label{eq:cma_loss}
\end{equation}
where $\lambda>0$ is an hyper-parameter that controls the weight of the two losses.
\paragraph{Implementation.} After Cross-\AVID pre-training, cross-modal disagreements are corrected by finetuning the audio and video networks to minimize the loss in~\cref{eq:cma_loss}. Models are initialized with the Cross-\AVID model at epoch 200, and trained for 200 additional epochs. We compare these models to a Cross-\AVID model trained for 400 epochs, thus controlling for the total number of parameter updates. For each sample, we find 32 positive instances using the \CMA criterion of~\cref{eqn:cma_sets} applied to video and audio memory bank representations. For efficiency purposes, the positive set is updated every 50 epochs. In each iteration, 1024 negative memories (not overlapping with positives) were sampled. These positive and negative memories were then used to minimize the CMA loss of Equations~\ref{eq:wm_nce}-\ref{eq:cma_loss}. For evaluation purposes, we use the same protocol as in~\cref{sec:analysis-avid}.

\subsection{Analyzing \CMA}
\label{sec:analysis-wmd}

\begin{figure}[t!]
    \centering
    \includegraphics[width=0.49\linewidth]{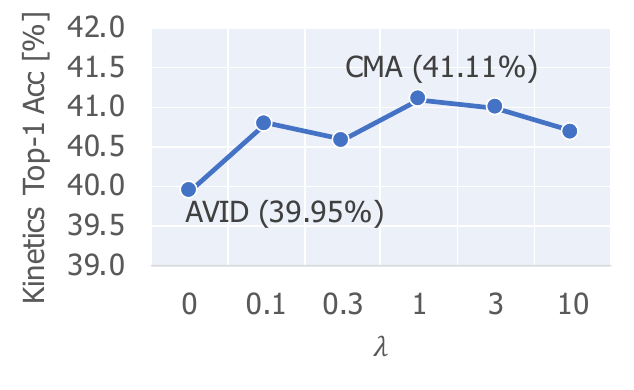} 
    \includegraphics[width=0.49\linewidth]{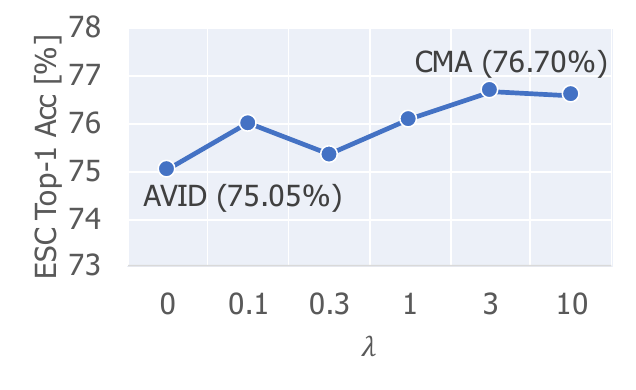}
    \caption{{\bf Ablation of \CMA objective.} Impact of within-modal positive sample discrimination. A network is pre-trained for different values of hyper-parameter $\lambda$ in \cref{eq:cma_loss}, and then evaluated by linear probing on the Kinetics and ESC datasets. Positive sample discrimination can further improve the performance of Cross-\AVID.}
    \label{fig:cma_lambda}
\end{figure}

The \CMA objective consists of two terms that optimize cross-modal (\cref{eq:cross_avid_loss}) and within-modal (\cref{eq:wm_nce}) similarity. We observed in~\cref{sec:analysis-avid} that within-modal comparisons for instance discrimination result in poor visual representations due to the relatively easy task of self-discrimination.
Intuitively, since \CMA identifies groups of instances ($\mathcal{P}_i$) that are likely related, calibrating within-modal similarity within these groups (instead of within the instance itself) should result in a better visual representation. 
To study this, we use \CMA to obtain a positive set $\mathcal{P}_i$ and analyse the CMA objective of \cref{eq:cma_loss} by evaluating with different values of the hyper-parameter $\lambda$. 
The results shown in~\cref{fig:cma_lambda} validates the advantages of \CMA over Cross-\AVID.

\paragraph{\CMA calibration.}
To understand the effect of the \CMA procedure on within-modal similarities, we analyzed the embedding space defined by memory bank representations obtained with \AVID and \CMA trained on the Kinetics dataset. 
Since representations are restricted to the unit sphere (due to normalization), the average inner-product between two randomly chosen samples should be 0 (assuming a uniform distribution of samples over the sphere). However, when training with Cross-\AVID, the average inner-product is 0.23. This means that Cross-AVID learns collapsed representations (\ie features are on average closer to other random features than the space permits). This is likely due to the lack of within-modal negatives when training for cross-modal discrimination. 
By seeking within modal-discrimination of positive samples, \CMA effectively addresses the feature collapsing problem observed for Cross-\AVID, and yields an average dot-product between random memories of 0 as expected.

\paragraph{\CMA \vs within-modal expansion.}
\CMA expands the positive set $\mathcal{P}_i$ to include instances that \emph{agree in both} video and audio spaces. We inspected whether modeling this agreement is necessary for relating instances by exploring alternatives that do not model agreements in both spaces (see~\cref{fig:CMA_vs_alternatives}). 
We consider alternatives that expand the set $\mathcal{P}_i$ by looking at instances that are similar in 1) only the audio space; 2) only the video space; or 3) either video or audio space.
Each method in~\cref{fig:CMA_vs_alternatives} is trained to optimize the objective of \cref{eq:cma_loss} with the corresponding $\mathcal{P}_i$. 
We also compare against the Cross-\AVID baseline that uses only the instance itself as the positive set.
Transfer performance is reported in~\cref{tab:analysis-cma}.

Compared to Cross-\AVID, expanding the set of positives using only audio similarity (third row) hurts performance on Kinetics, and relying on video similarities alone (second row) only provides marginal improvements.
We believe that expanding the set of positives only based on visual similarity does not improve the performance of visual features since the positives are \emph{already close} in the feature space, and do not add extra information.
\CMA provides consistent gains over all methods on Kinetics, suggesting that modeling \emph{agreement} can provide better positive sets for representation learning of visual features.

\newcommand{\VideoKineticsCMATable}{
\begin{tabular}{l|cccc|c}
    \toprule
    \bf Method & \bf block1 & \bf block2 & \bf block3 & \bf block4 & \bf Best \\
    \hline\hline
    \rcg Cross-\AVID (Base) & 19.80 & 26.98 & 34.81 & 39.95 & 39.95 \\
    \rcw Base + Video-Exp.  & 19.93 & 27.39 & 35.64 & 40.17 & 40.17 \\
    \rcg Base + Audio-Exp.  & 20.14 & 27.28 & 35.68 & 39.62 & 39.62 \\
    \rcw Base + AV Exp      & 20.04 & 27.61 & 36.14 & 40.58 & 40.58 \\
    \rcg Base + CMA         & \bf 20.16 & \bf 27.98 & \bf 36.98 & \bf 41.11 & \bf 41.11 \\
    \bottomrule
    \mc{4}{c}{}
\end{tabular}
}

\begin{figure*}[t!]
    \centering
    \begin{subfigure}{0.4\textwidth}
        \includegraphics[width=0.95\linewidth]{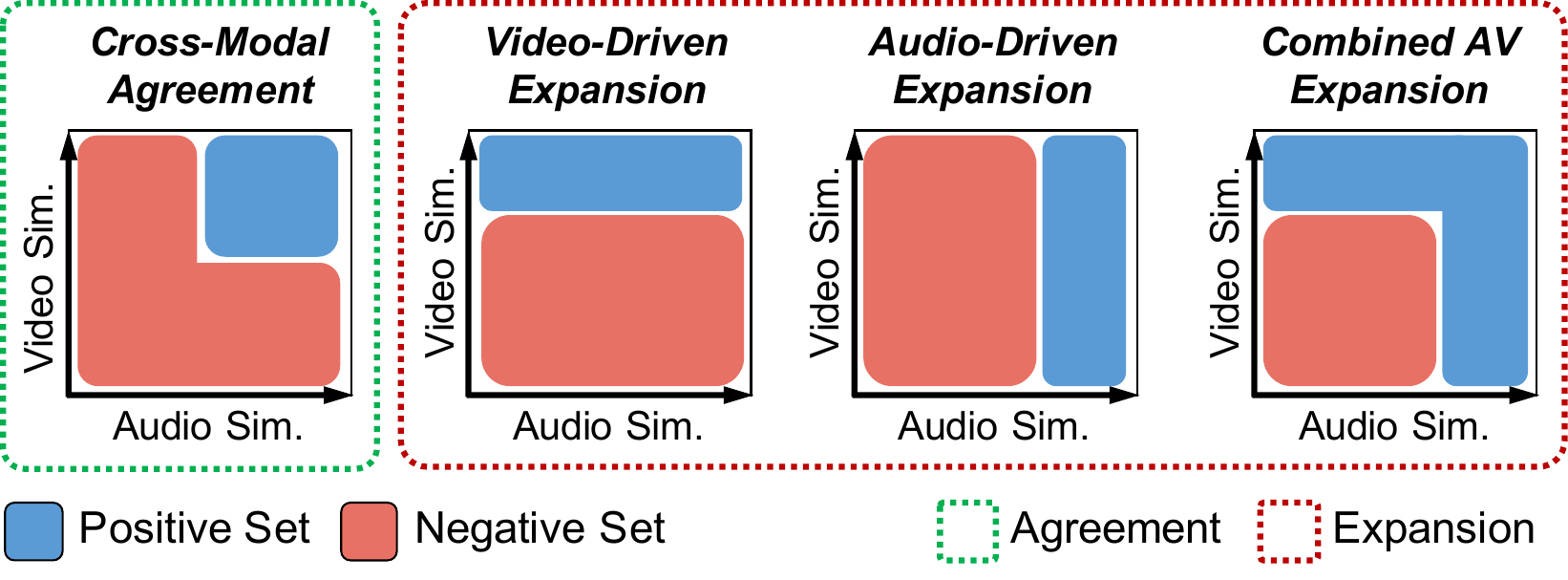} 
        \caption{Positive and negative sets of `agreement' and `expansion' methods.}
        \label{fig:CMA_vs_alternatives}
    \end{subfigure}\;
    \begin{subfigure}{0.37\textwidth}
        \resizebox{\linewidth}{!}{\VideoKineticsCMATable}
        \caption{Top-1 accuracy of linear probing on Kinetics.\label{tab:analysis-cma}}
    \end{subfigure}\;
    \begin{subfigure}{0.2\textwidth}
        \includegraphics[width=\linewidth]{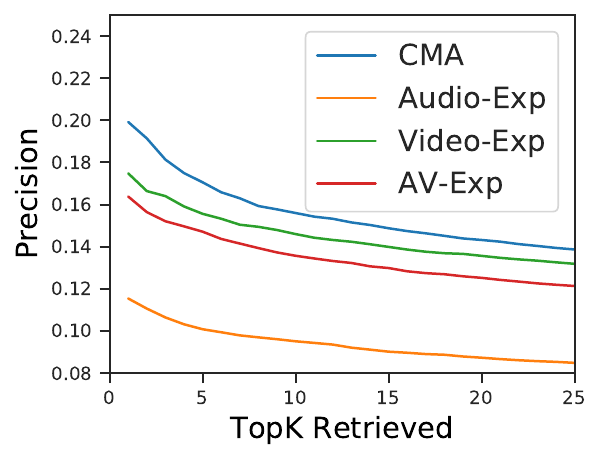}
        \caption{Precision@K.}
        \label{fig:cma_precision}
    \end{subfigure}
    \caption{\textbf{\CMAName \vs. Within-modality Expansion} We study the importance of modeling \emph{agreement} between video and audio similarities. We compare CMA to expansion methods that relate instances without modeling agreement  (\ref{fig:CMA_vs_alternatives}). CMA enables better transfer for action recognition (\ref{tab:analysis-cma}). Expansion methods generate agreements of worse precision (\ref{fig:cma_precision}).}
\end{figure*}

\begin{figure}[t!]
    \centering
    \includegraphics[width=\linewidth]{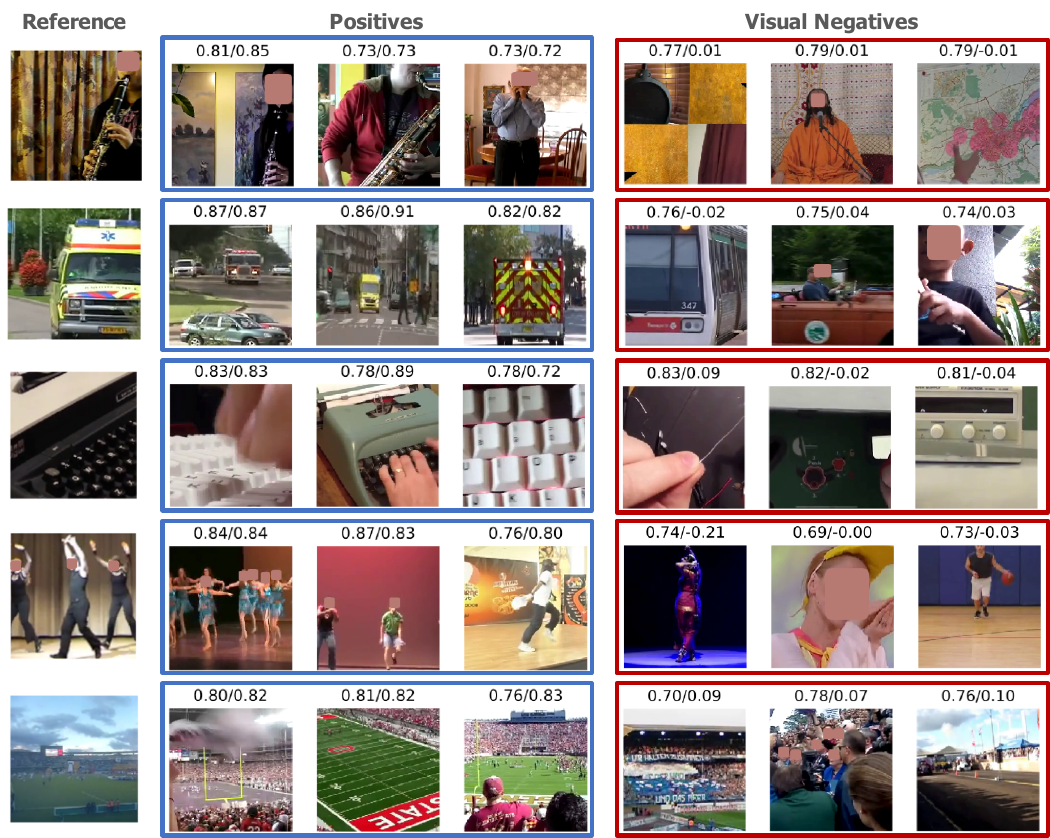}
    \caption{\textbf{Examples extracted by the \CMA procedure.} For each reference image, we show three images in their positive sets (\cref{eqn:cma_sets}). We also show three negatives that were rejected from the positive set due to low audio similarity. Each image is annotated with the video/audio similarity to the reference.}
    \label{fig:positives}
\end{figure}

\paragraph{Qualitative Understanding.} 
We show examples of positive and negative samples found by \CMA in~\cref{fig:positives} and observe that \CMA can group together semantically related concepts. As it uses agreement between both spaces, visually similar concepts, like `ambulance` and `bus` (second row), can be distinguished based on audio similarity. 
This leads to more precise positive sets $\mathcal{P}_i$, as can be verified by inspecting the precision$@K$ of $\mathcal{P}_i$ measured against ground truth labels (\cref{fig:cma_precision}).
\CMA consistently finds more precise positives compared to within-modal expansion methods showing the advantages of modeling agreement.

\section{Cross-AVID and CMA at scale}
\label{sec:experiments}

Previous sections provide experimental validation for the proposed Cross-AVID and CMA procedures when training on a medium-sized dataset (100K videos from Audioset). We now study the proposed methods on large-scale datasets. We also compare Cross-AVID and CMA to prior work, including video-based self-supervised learning methods~\cite{shuffle, opn, xu2019self, dpc}, and methods that leverage the natural correspondence between audio and video~\cite{l3,owens,bruno_avts}.


\paragraph{Experimental setup.}
We briefly describe the experimental setup, and refer the reader to supplementary material for full details.
We use the 18-layer R(2+1)D network of~\cite{tran2018closer} as the video encoder and a 9-layer (2D) CNN with batch normalization as the audio encoder. 
Models are trained on Kinetics-400~\cite{kinetics} and the full Audioset~\cite{audioset} datasets, containing 240K and 1.8M video instances, respectively.
Video clips composed of $8$ frames of size $224\!\times\!224$ are extracted at a frame rate of 16fps with standard data augmentation
procedures~\cite{szegedy2015going}.
Two seconds of audio is randomly sampled within 0.5 seconds of the video at a 24kHz sampling rate, and spectrograms of size $200\times257$ (200 time steps with 257 frequency bands) are used as the input to the audio network.
For Cross-\AVID, the cross-modal discrimination loss of Equation~\ref{eq:cross_avid_loss} is optimized with $K=1024$ negative instances. We then find 128 positive instances for each sample using cross-modal agreements (\cref{eqn:cma_sets}), and optimize the \CMA criterion of \cref{eq:cma_loss} with $K_p=32$ positives, $K_n=1024$ negatives and $\lambda=1.0$. 
Video representations are evaluated on action recognition (\cref{sec:action}), and audio representations on sound classification (\cref{sec:audio}).

\subsection{Action recognition}
\label{sec:action}

We first evaluate the visual representations learned by Cross-\AVID and AVID+\CMA by training a linear classifier for the task of action recognition on the Kinetics dataset. The top-1 accuracy is reported for clip and video-level predictions. Clip-level predictions are obtained from a single 8-frame clip, while video-level predictions are computed by averaging clip-level predictions from 10 clips uniformly sampled from the whole video.
The results shown in \cref{tab:kinetics} clearly demonstrate the advantage of calibrating AVID representations using the CMA procedure, yielding significant gains across both metrics and pretraining datasets. These results demonstrate the value of the CMA procedure in large-scale datasets, thus showing that its effect goes beyond a simple regularization procedure to prevent overfitting.

\begin{table}[t!]
    \centering
    \resizebox{0.9\linewidth}{!}{
    \begin{tabular}{cp{2mm}ccp{2mm}cc}
        \toprule
        \bf Pretraining DB && \mc{2}{c}{\bf Kinetics} && \mc{2}{c}{\bf Audioset} \\
        \bf Method \textbackslash\; Metric && Clip@1 & Video@1 && Clip@1 & Video@1 \\
        \hline\hline
        \rcg Cross-AVID && 33.3 & 43.1 && 35.2 & 46.6 \\
        \rcw AVID+CMA    && \bf 35.1 & \bf 44.5 && \bf 37.4 & \bf 48.9 \\
        \bottomrule
    \end{tabular}}
    \caption{Top-1 accuracy of linear probing on Kinetics.}
    \label{tab:kinetics}
\end{table}

To compare to prior work, we follow~\cite{dpc,bruno_avts,cmc} and evaluate visual representations on the UCF-101~\cite{ucf} and HMDB-51~\cite{hmdb} datasets, by full network fine-tuning. Due to the large variability of experimental setups used in the literature, it is unrealistic to provide a direct comparison to all methods, as these often use different network encoders trained on different datasets with input clips of different lengths. To increase the range of meaningful comparisons, we fine-tuned our models using clips with both 8 and 32 frames.
At inference time, video-level predictions are provided by averaging clip-level predictions for 10 uniformly sampled clips~\cite{bruno_avts}.
We report top-1 accuracy averaged over the three train/test splits provided with the original datasets.

\newcommand{\ActionSOTAbyDB}{
\begin{tabular}{rcccc}
\toprule
\bf Method & \bf \begin{tabular}{c}Pretraining\\DB\end{tabular} & \bf \begin{tabular}{c}Finetune \\Input Size\end{tabular} & \bf UCF & \bf HMDB \\
\hline\hline
\rcg Shuffle\&Learn~\cite{shuffle} & UCF & $1\!\times\!227^2$ & 50.2 & 18.1 \\
\rcw OPN~\cite{opn}                & UCF & $1\!\times\!227^2$ & 56.3 & 23.8 \\
\rcg ST Order~\cite{storder}       & UCF & $1\!\times\!227^2$ & 58.6 & 25.0 \\
\rcw CMC~\cite{cmc}                & UCF & $1\!\times\!227^2$ & 59.1 & 26.7 \\ 
\hline\hline
\rcg 3D-RotNet~\cite{3d_rotation}  & Kinetics400 & $16\!\times\!112^2$ & 62.9 & 33.7 \\
\rcw ClipOrder~\cite{xu2019self}   & Kinetics400 & $16\!\times\!112^2$ & 72.4 & 30.9 \\
\rcg DPC~\cite{dpc}                & Kinetics400 & $25\!\times\!128^2$ & 75.7 & 35.7 \\
\rcw CBT~\cite{cbt}                & Kinetics400 & $16\!\times\!112^2$ & 79.5 & 44.6 \\
\rcg L3${}^*$~\cite{l3}            & Kinetics400 & $16\!\times\!224^2$ & 74.4 & 47.8 \\ 
\rcw AVTS~\cite{bruno_avts}        & Kinetics400 & $25\!\times\!224^2$ & 85.8 & 56.9 \\
\rcg                               & Kinetics400 & $8\!\times\!224^2$  & 74.2 & 39.0 \\ 
\rcg \mr{-2}{XDC~\cite{xmodalclust}} & Kinetics400 & $32\!\times\!224^2$ & 86.8${}^\dagger$ & 52.6${}^\dagger$ \\
\rcw                               & Kinetics400 & $8\!\times\!224^2$  & 82.3 & 49.1 \\
\rcw \mr{-2}{Cross-\AVID (ours)}   & Kinetics400 & $32\!\times\!224^2$ & \uline{86.9} & \uline{59.9} \\
\rcg                               & Kinetics400 & $8\!\times\!224^2$  & 83.7 & 49.5 \\
\rcg \mr{-2}{AVID+CMA (ours)}      & Kinetics400 & $32\!\times\!224^2$ & \bf 87.5 & \bf 60.8 \\
\hline\hline
\rcw L3${}^*$~\cite{l3}            & Audioset & $16\!\times\!224^2$ & 82.3 & 51.6 \\ 
\rcg Multisensory~\cite{owens}     & Audioset & $64\!\times\!224^2$ & 82.1 & --   \\ 
\rcw AVTS~\cite{bruno_avts}        & Audioset & $25\!\times\!224^2$ & 89.0 & 61.6 \\
\rcg                               & Audioset & $8\!\times\!224^2$  & 84.9 & 48.8 \\
\rcg \mr{-2}{XDC~\cite{xmodalclust}} & Audioset & $32\!\times\!224^2$ & 93.0${}^\dagger$ & 63.7${}^\dagger$ \\
\rcw                          & Audioset & $8\!\times\!224^2$     & 88.3 & 57.5 \\
\rcw \mr{-2}{Cross-\AVID (ours)} & Audioset & $32\!\times\!224^2$ & \uline{91.0} & \uline{64.1} \\
\rcg                          & Audioset & $8\!\times\!224^2$     & 88.6 & 57.6 \\
\rcg \mr{-2}{AVID+CMA (ours)} & Audioset & $32\!\times\!224^2$    & \bf 91.5 & \bf 64.7 \\
\bottomrule 
\end{tabular}
}
\begin{table}[t!]
    \centering
    \resizebox{\columnwidth}{!}{\ActionSOTAbyDB}
    \caption{Top-1 accuracy on UCF and HMDB  by full network finetuning with various pre-training datasets and clips of different sizes. Methods were organized by pre-training dataset. The method with the best performance is indicated in bold face, and second best is underlined. ${}^*$Re-implemented by us. ${}^\dagger$Obtained by pre-training and finetuning with larger $32\times224^2$ inputs (we only pre-train on $8\times224^2$ inputs).\label{tab:action-sota}}
\end{table}

Table~\ref{tab:action-sota} compares the transfer performance of Cross-\AVID and \CMA with previous self-supervised approaches.
To enable well-grounded comparisons, we also list for each method the pre-training dataset and clip dimensions used while finetuning on UCF and HMDB. Despite its simplicity, Cross-\AVID achieves state-of-the-art performance for equivalent data settings in most cases. In particular, when pre-trained on Audioset, Cross-\AVID outperformed other audio-visual SSL methods such as L3 and AVTS by at least 1.0\% on UCF and 2.5\% on HMDB. Similar to Cross-\AVID, L3 and AVTS propose to learn audio-visual representations by predicting whether audio/video pairs are in-sync. However, these methods optimize for the audiovisual correspondence task, which fails to reason about the data distribution at large.
Cross-\AVID also outperformed the concurrently proposed XDC~\cite{xmodalclust} under equivalent data settings. 
When pretrained on Audioset and finetuned on UCF with 32 frames, XDC~\cite{xmodalclust} does report higher accuracy, but the model was pretrained and finetuned using 32 frames, while we pretrain using only 8 frames. It should be noted that, when pretraining and finetuning with clips of 8 frames, Cross-AVID outperforms XDC by 3.4\% (84.9\% vs 88.3\%).
\CMA further improves the performance of Cross-\AVID on all settings considered (\ie, using both Kinetics and Audioset pretraining datasets, and evaluating on UCF and HMDB).
We observed, however, that the improvements of \CMA over Cross-\AVID are smaller under the fine-tuning protocol than the linear evaluation of~\cref{tab:kinetics}. 
Prior work~\cite{scaling,splitbrain} observes that full fine-tuning significantly modifies the visual features and tests the network initialization aspect of pre-training rather than the semantic quality of the representation. Thus, we believe that the feature calibration benefits of \CMA are diminished under the full finetuning protocol.

\subsection{Sound recognition}
\label{sec:audio}

Audio representations are evaluated on the ESC-50~\cite{esc} and DCASE~\cite{dcase} datasets by linear probing~\cite{scaling} for the task of sound recognition. 
Following~\cite{bruno_avts}, both ESC and DCASE results are obtained by training a linear one-vs-all SVM classifier on the audio representations generated by the pre-trained models at the final layer before pooling.
For training, we extract 10 clips per sample on the ESC dataset and 60 clips per sample on DCASE~\cite{bruno_avts}.
At test time, sample level predictions are obtained by averaging 10 clip level predictions, and the top-1 accuracy is reported in \cref{tab:audio-sota}.
For the ESC dataset, performance is the average over the 5 original train/test splits. 
Similarly to video, audio representations learned by Cross-\AVID and \CMA outperform prior work, outperforming ConvRBM on the ESC dataset by 2.7\% and AVTS on DCASE by 3\%.

\newcommand{\AudioSOTAbyDB}{
\begin{tabular}{rccc}
\toprule
\bf Method                              & \bf \begin{tabular}{c}Pretraining\\DB\end{tabular} & \bf ESC & \bf DCASE \\ 
\hline\hline
\rcg RandomForest~\cite{esc}         & None & 44.3 & -- \\
\rcw ConvNet~\cite{piczakconvnet}    & None & 64.5 & -- \\
\rcg ConvRBM~\cite{convrbm}          & None & 86.5 & -- \\
\hline\hline
\rcw SoundNet~\cite{soundnet}        & Flickr-SoundNet & 74.2 & 88 \\
\rcg L3~\cite{l3}                    & Flickr-SoundNet & 79.3 & 93 \\ 
\hline\hline
\rcw AVTS~\cite{bruno_avts}   & Kinetics & 76.7 & 91 \\ 
\rcg XDC~\cite{xmodalclust}   & Kinetics & \uline{78.5} & -- \\
\rcw Cross-\AVID (Ours)       & Kinetics &77.6 & \bf \uline{93} \\
\rcg AVID+CMA (Ours)          & Kinetics &\bf 79.1 & \bf \uline{93} \\
\hline\hline
\rcw AVTS~\cite{bruno_avts} & Audioset & 80.6 & 93 \\
\rcg XDC~\cite{xmodalclust} & Audioset & 85.8 & -- \\
\rcw Cross-\AVID (Ours)     & Audioset & \bf 89.2 & \bf \uline{96} \\
\rcg AVID+CMA (Ours)        & Audioset & \uline{89.1} & \bf \uline{96} \\
\bottomrule 
\end{tabular}
}

\begin{table}[t!]
    \centering
    \resizebox{0.8\columnwidth}{!}{\AudioSOTAbyDB}
    \caption{Top-1 accuracy of linear classification on ESC-50 and DCASE datasets. Methods are organized by pre-training dataset. The method with the best performance is indicated in bold face, and second best is underlined. \label{tab:audio-sota}}
\end{table}

\section{Discussion}
We proposed a self-supervised method to learn visual and audio representations by contrasting visual representations against multiple audios, and vice versa. Our method, Audio-Visual Instance Discrimination (\AVID) builds upon recent advances in contrastive learning~\cite{instance,cmc} to learn state-of-the-art representations that outperform prior work on action recognition and sound classification. We propose and analyze multiple variants of the \AVID task to show that optimizing for cross-modal similarity and not within-modal similarity matters for learning from video and audio.

We also identified key limitations of the instance discrimination framework and proposed \CMA to use agreement in the video and audio feature spaces to group together related videos. \CMA helps us relate \emph{multiple} instances by identifying more related videos. \CMA also helps us reject `false positives', \ie, videos that are similar visually but differ in the audio space. We show that using these groups of related videos allows us to optimize for within-modal similarity, in addition to cross-modal similarity, and improve visual and audio representations. The generalization of \CMA suggests that cross-modal agreements provide non-trivial correspondences between samples and are a useful way to learn improved representations in a multi-modal setting.

\section*{Acknowledgements}
We are grateful to Rob Fergus and Laurens van der Maaten for their feedback and support; Rohit Girdhar for feedback on the manuscript; and Bruno Korbar for help with the baselines.

{\small
\bibliographystyle{ieee_fullname}
\bibliography{refs}
}

\vfill\pagebreak\ 
\vfill\pagebreak
\appendix

\section{Experimental setup}
\paragraph{Architecture details}
The architecture details of the video and audio networks used in the analysis experiments are shown in~\cref{tab:arch-analysis}, and those used for comparison to prior work is shown in~\cref{tab:arch-sota}.

\paragraph{Pre-training hyper-parameters}
Optimization and data augmentation hyper-parameters for AVID and CMA pre-training are provided in~\cref{tab:pretrain-params}.

\paragraph{Action recognition hyper-parameters}
Optimization and data augmentation hyper-parameters for action recognition tasks are provided in~\cref{tab:transfer-params}.

\paragraph{Video pre-processing}
Video clips are extracted at 16 fps and augmented with standard techniques, namely random multi-scale cropping with 8\% minimum area, random horizontal flipping and color and temporal jittering. Color jittering hyper-parameters are shown in \cref{tab:pretrain-params} for pre-training and \cref{tab:transfer-params} for transfer into downstream tasks. 

\paragraph{Audio pre-processing}
Audio signals are loaded at 24kHz, instead of 48kHz, because a large number of Audioset audio samples do not contain these high frequencies. The spectrogram is computed by taking the FFT on 20ms windows with either 10ms (\S4, \S5) or 20ms (\S6) hop-size.
We then convert the spectrogram to a log scale, and Z-normalize its intensity using mean and standard deviation values computed on the training set. We use volume and temporal jitering for data augmentation. Volume jittering is accomplished by multiplying the audio waveform by a constant factor randomly sampled between 0.9 and 1.1, and applied uniformly over time. Temporal jittering is done by randomly sampling the audio starting time within 0.5s of the video, and randomly selecting the total audio duration between 1.4s and 2.8s and rescaling back to the expected number of audio frames.

\section{Longer \AVID pre-training}
To ensure that the benefits of \CMA are not caused by longer training, we trained Cross-\AVID for the same number of epochs as AVID+\CMA. The Cross-\AVID performance on Kinetics after 200 and 400 training epochs are shown in~\cref{tab:avid-longer}.
Cross-\AVID transfer performance seem to have already saturated after 200 epochs of pre-training.

\section{CMA calibration}
To further study the benefits effect of the CMA procedure, we measured the classification performance of memory representations obtained with both AVID and CMA trained on the Kinetics dataset. We randomly split the 220K training samples, for which memory representations are available, into a train/validation set (70/30\% ratio). We then train a linear classifier on the training set (using either video, audio or the concatenation of both, ConvNet is kept fixed), and evaluate the performance on the validation set. The train/validation splits are sampled 5 times and average performance is reported. The top-1 accuracies are shown in \cref{tab:mem-cls-calibration}. 

\begin{table}[h!]
\caption{Top-1 accuracy of linear probing on Kinetics evaluated after 200 and 400 epochs of Cross-\AVID training. \label{tab:avid-longer} }
\centering
\resizebox{\linewidth}{!}{
\begin{tabular}{|r|cccc|c|}
\hline
\bf Method & \bf block1 & \bf block2 & \bf block3 & \bf block4 & \bf Best \\ \hline
Cross-AVID (ep 200) & 19.84 & 26.87 & 34.64 & 39.87 & 39.87 \\
Cross-AVID (ep 400) & 19.80 & 26.98 & 34.81 & 39.95 & 39.95 \\
\hline
\end{tabular}}
\end{table}

\begin{table}[h!]
\centering
\caption{Top-1 accuracy of linear probing of memory representations (video, audio and both concatenated). \label{tab:mem-cls-calibration}}
\resizebox{\linewidth}{!}{
\begin{tabular}{|r|ccc|}
\hline
\bf Method & \bf Video Mem & \bf Audio Mem  &  \bf Combined Mem \\ \hline
\bf Cross-AVID   & 29.01$\pm$0.14 & 19.67$\pm$0.09 & 34.68$\pm$0.15 \\
\bf CMA    & \bf 34.00$\pm$0.25 & \bf 21.98$\pm$0.11 & \bf 38.91$\pm$0.14 \\
\hline
\end{tabular}}
\end{table}

\begin{table}[h!]
\caption{Pre-training optimization hyper-parameters. CMA models are initialized by the AVID model obtained at epoch 200. bs batch size; lr learning rate; wd weight decay; ep number of epochs; es number of samples per epoch; msc - multi-scale cropping; hf - horizontal flip probability; bj/sj/cj/hj - brightness/saturation/contrast/hue jittering intensity.}
\label{tab:pretrain-params}
\centering
\resizebox{\linewidth}{!}{
\begin{tabular}{|r|c|ccccc|cccccc|}
\hline
\bf Method & \bf DB & \bf bs & \bf lr   & \bf wd   & \bf ep  & \bf es & \bf msc & \bf hf & \bf bj & \bf sj & \bf cj & \bf hj \\\hline 
\bf AVID (\S4) & Audioset & 32 & 5e-4 & 1e-5 & 400 & 1e5          & \checkmark & 0.5 & 0.4 & 0.4 & 0.4 & 0.2 \\
\bf AVID (\S6) & Audioset & 32 & 5e-4 & 1e-5 & 200 & 1.8e6        & \checkmark & 0.5 & 0.4 & 0.4 & 0.4 & 0.2 \\
\bf AVID (\S6) & Kinetics & 32 & 2e-4 & 1e-5 & 300 & 2.4e5        & \checkmark & 0.5 & 0.4 & 0.4 & 0.4 & 0.2 \\\hline
\bf CMA (\S5.3, \S5.4) & Audioset & 32 & 5e-4 & 1e-5 & 200 & 1e5  & \checkmark & 0.5 & 0.4 & 0.4 & 0.4 & 0.2 \\
\bf CMA (\S6)  & Audioset & 32 & 5e-4 & 1e-5 & 200 & 1.8e6        & \checkmark & 0.5 & 0.4 & 0.4 & 0.4 & 0.2\\
\bf CMA (\S6)  & Kinetics & 32 & 2e-4 & 1e-5 & 300 & 2.4e5        & \checkmark & 0.5 & 0.4 & 0.4 & 0.4 & 0.2\\ 
\hline
\end{tabular}}
\end{table}

\begin{table}[h!]
\caption{Transfer learning optimization and data augmentation hyper-parameters. bs - batch size; lr - learning rate; wd - weight decay; ep - number of epochs; es - number of samples per epoch; gm - learning rate decay factor; mls - milestones for learning rate decay; msc - multi-scale cropping; hf - horizontal flip probability; bj/sj/cj/hj - brightness/saturation/contrast/hue jittering intensity.}
\label{tab:transfer-params}
\centering
\resizebox{\linewidth}{!}{
\begin{tabular}{|r|c|ccccccc|}
\hline
\bf DB   & \bf input size & \bf bs & \bf lr   & \bf wd   & \bf ep  & \bf es & \bf gm & \bf mls \\\hline
\bf Kinetics (\S4, \S5) & $16\times112^2$ & 32 & 1e-4 & 0. & 20 & 1e4 & 0.3 & 8,12,15,18 \\\hline
\bf UCF (\S6) & $8\times224^2$  & 32 & 1e-4 & 0. & 160 & 1e4 & 0.3 & 60,100,140 \\
\bf UCF (\S6) & $32\times224^2$ & 16 & 1e-4 & 0. & 80  & 1e4 & 0.3 & 30,50,70 \\ \hline
\bf HMDB (\S6) & $8\times224^2$  & 32 & 1e-4 & 0. & 250 & 3.4e3 & 0.3 & 75,150,200 \\
\bf HMDB (\S6) & $32\times224^2$ & 16 & 1e-4 & 0. & 100 & 3.4e3 & 0.3 & 30,60,80 \\
\hline
\multicolumn{9}{c}{}\\
\end{tabular}}
\resizebox{0.6\linewidth}{!}{
\begin{tabular}{|r|cccccc|}
\hline
\bf DB   & \bf msc & \bf hf & \bf bj & \bf sj & \bf cj & \bf hj \\\hline
\bf Kinetics (\S4, \S5) & \checkmark & 0.5 & 0. & 0. & 0. & 0. \\
\bf UCF (\S6)  & \checkmark & 0.5 & 0.4 & 0.4 & 0.4 & 0.2  \\
\bf HMDB (\S6) & \checkmark & 0.5 & 1.  & 1.  & 1.  & 0.2 \\
\hline
\end{tabular}}
\end{table}
\begin{table}[th!]
\caption{Architecture details of R(2+1)D video network and Conv2D audio network for analysis experiments (\S{}4, \S{}5.3, \S{}5.4). The video network is based of R(2+1)D convolutions, and the audio on 2D convolutions. Both video and audio networks use ReLU activations and batch normalization at each layer. $X_s$ spatial activation size, $X_t$ temporal activation size, $X_f$ frequency activation size, $C$ number of channels, $K_s$ spatial kernel size, $K_t$ temporal kernel size, $K_f$ frequency kernel size, $S_s$ spatial stride, $S_t$ temporal stride, $S_f$ frequency stride.}
\label{tab:arch-analysis}
\centering
\resizebox{0.7\linewidth}{!}{
\begin{tabular}{|r|ccccccc|}
\mc{8}{c}{}\\
\mc{8}{c}{\bf Video Network}\\
\hline
\bf Layer    & \bf $X_s$ & \bf $X_t$ & \bf $C$ & \bf $K_s$ & \bf $K_t$ & \bf $S_s$ & \bf $S_t$ \\\hline
\bf \texttt{video}    & 112 & 16    & 3   & -     & -     & -     & -     \\
\bf \texttt{conv1}    & 56  & 16    & 64  & 7     & 3     & 2     & 1     \\
\bf \texttt{block2.1}   & 56  & 16    & 64  & 3     & 3     & 1     & 1     \\
\bf \texttt{block2.2}   & 56  & 16    & 64  & 3     & 3     & 1     & 1     \\
\bf \texttt{block3.1}   & 28  & 8     & 128 & 3     & 3     & 2     & 2     \\
\bf \texttt{block3.2}   & 28  & 8     & 128 & 3     & 3     & 1     & 1     \\
\bf \texttt{block4.1}   & 14  & 4     & 256 & 3     & 3     & 2     & 2     \\
\bf \texttt{block4.2}   & 14  & 4     & 256 & 3     & 3     & 1     & 1     \\
\bf \texttt{block5.1}   & 7   & 2     & 512 & 3     & 3     & 2     & 2     \\
\bf \texttt{block5.2}   & 7   & 2     & 512 & 3     & 3     & 1     & 1     \\
\bf \texttt{max pool} & 1   & 1     & 512 & 7     & 2     & 1     & 1     \\
\bf \texttt{fc1}       & -   & -     & 512 & -     & -     & -     & -    \\
\bf \texttt{fc2}       & -   & -     & 512 & -     & -     & -     & -    \\
\bf \texttt{fc3}       & -   & -     & 128 & -     & -     & -     & -    \\
\hline
\multicolumn{8}{!}{}\\
\end{tabular}}
\resizebox{0.7\linewidth}{!}{
\begin{tabular}{|r|ccccccc|}
\mc{8}{c}{}\\
\mc{8}{c}{\bf Audio Network}\\
\hline
\bf Layer      & \bf $X_f$ & \bf $X_t$ & \bf $C$ & \bf $K_f$ & \bf $K_t$ & \bf $S_f$ & \bf $S_t$ \\ \hline
\bf \texttt{audio}    & 129 & 100 & 1 & - & - & - & - \\
\bf \texttt{conv1}    & 65 & 50 & 64 & 7 & 7 & 2 & 2 \\
\bf \texttt{block2.1}   & 65 & 50 & 64 & 3 & 3 & 1 & 1 \\
\bf \texttt{block2.2}   & 65 & 50 & 64 & 3 & 3 & 1 & 1 \\
\bf \texttt{block3.1}   & 33 & 25 & 128 & 3 & 3 & 2 & 2 \\
\bf \texttt{block3.2}   & 33 & 25 & 128 & 3 & 3 & 1 & 1 \\
\bf \texttt{block4.1}   & 17 & 13 & 256 & 3 & 3 & 2 & 2 \\
\bf \texttt{block4.2}   & 17 & 13 & 256 & 3 & 3 & 1 & 1 \\
\bf \texttt{block5.1}   & 17 & 13 & 512 & 3 & 3 & 1 & 1 \\
\bf \texttt{block5.2}   & 17 & 13 & 512 & 3 & 3 & 1 & 1 \\
\bf \texttt{max pool} & 1 & 1 & 512 & 17 & 13 & 1 & 1 \\
\bf \texttt{fc1}       & - & - & 512 & - & - & - & - \\ 
\bf \texttt{fc2}       & - & - & 512 & - & - & - & - \\ 
\bf \texttt{fc3}       & - & - & 128 & - & - & - & - \\ 
\hline
\end{tabular}
}
\end{table}

\begin{table}[th!]
\caption{Architecture details of R(2+1)D video network and Conv2D audio network for comparison to prior work (\S{}6).  The video network is based of R(2+1)D convolutions, and the audio on 2D convolutions. Both video and audio networks use ReLU activations and batch normalization at each layer. $X_s$ spatial activation size, $X_t$ temporal activation size, $X_f$ frequency activation size, $C$ number of channels, $K_s$ spatial kernel size, $K_t$ temporal kernel size, $K_f$ frequency kernel size, $S_s$ spatial stride, $S_t$ temporal stride, $S_f$ frequency stride.}
\label{tab:arch-sota}
\centering
\resizebox{0.7\linewidth}{!}{
\begin{tabular}{|r|ccccccc|}
\mc{8}{c}{}\\
\mc{8}{c}{\bf Video Network}\\
\hline
\bf Layer    & \bf $X_s$ & \bf $X_t$ & \bf $C$ & \bf $K_s$ & \bf $K_t$ & \bf $S_s$ & \bf $S_t$ \\\hline
\bf \texttt{video}    & 224 & 8 & 3   & -  & - & - & - \\
\bf \texttt{conv1}    & 112 & 8 & 64  & 7  & 3 & 2 & 1 \\
\bf \texttt{max-pool} & 56  & 8 & 64  & 3  & 1 & 2 & 1 \\
\bf \texttt{block2.1.1} & 56  & 8 & 64  & 3  & 3 & 1 & 1 \\
\bf \texttt{block2.1.2} & 56  & 8 & 64  & 3  & 3 & 1 & 1 \\
\bf \texttt{block2.2.1} & 56  & 8 & 64  & 3  & 3 & 1 & 1 \\
\bf \texttt{block2.2.2} & 56  & 8 & 64  & 3  & 3 & 1 & 1 \\
\bf \texttt{block3.1.1} & 28  & 4 & 128 & 3  & 3 & 2 & 2 \\
\bf \texttt{block3.1.2} & 28  & 4 & 128 & 3  & 3 & 1 & 1 \\
\bf \texttt{block3.2.1} & 28  & 4 & 128 & 3  & 3 & 1 & 1 \\
\bf \texttt{block3.2.2} & 28  & 4 & 128 & 3  & 3 & 1 & 1 \\
\bf \texttt{block4.1.1} & 14  & 2 & 256 & 3  & 3 & 2 & 2 \\
\bf \texttt{block4.1.2} & 14  & 2 & 256 & 3  & 3 & 1 & 1 \\
\bf \texttt{block4.2.1} & 14  & 2 & 256 & 3  & 3 & 1 & 1 \\
\bf \texttt{block4.2.2} & 14  & 2 & 256 & 3  & 3 & 1 & 1 \\
\bf \texttt{block5.1.1} & 7   & 1 & 512 & 3  & 3 & 2 & 2 \\
\bf \texttt{block5.1.2} & 7   & 1 & 512 & 3  & 3 & 1 & 1 \\
\bf \texttt{block5.2.1} & 7   & 1 & 512 & 3  & 3 & 1 & 1 \\
\bf \texttt{block5.2.2} & 7   & 1 & 512 & 3  & 3 & 1 & 1 \\
\bf \texttt{max-pool} & 1   & 1 & 512 & 7  & 2 & 1 & 1 \\
\bf \texttt{fc1}      & -   & - & 512 & -  & - & - & - \\
\bf \texttt{fc2}      & -   & - & 512 & -  & - & - & - \\
\bf \texttt{fc3}      & -   & - & 128 & -  & - & - & - \\
\hline
\multicolumn{8}{!}{}\\
\end{tabular}}
\resizebox{0.7\linewidth}{!}{
\begin{tabular}{|r|ccccccc|}
\mc{8}{c}{}\\
\mc{8}{c}{\bf Audio Network}\\
\hline
\bf Layer      & \bf $X_f$ & \bf $X_t$ & \bf $C$ & \bf $K_f$ & \bf $K_t$ & \bf $S_f$ & \bf $S_t$ \\ \hline
\bf \texttt{audio}    & 257 & 200 & 1 & - & - & - & - \\
\bf \texttt{conv1}    & 129 & 100 & 64 & 7 & 7 & 2 & 2 \\
\bf \texttt{block2.1}   & 65 & 50 & 64 & 3 & 3 & 2 & 2 \\
\bf \texttt{block2.2}   & 65 & 50 & 64 & 3 & 3 & 1 & 1 \\
\bf \texttt{block3.1}   & 33 & 25 & 128 & 3 & 3 & 2 & 2 \\
\bf \texttt{block3.2}   & 33 & 25 & 128 & 3 & 3 & 1 & 1 \\
\bf \texttt{block4.1}   & 17 & 13 & 256 & 3 & 3 & 2 & 2 \\
\bf \texttt{block4.2}   & 17 & 13 & 256 & 3 & 3 & 1 & 1 \\
\bf \texttt{block5.1}   & 17 & 13 & 512 & 3 & 3 & 1 & 1 \\
\bf \texttt{block5.2}   & 17 & 13 & 512 & 3 & 3 & 1 & 1 \\
\bf \texttt{max pool}   & 1 & 1 & 512 & 17 & 13 & 1 & 1 \\
\bf \texttt{fc1}        & - & - & 512 & - & - & - & - \\
\bf \texttt{fc2}        & - & - & 512 & - & - & - & - \\
\bf \texttt{fc3}        & - & - & 128 & - & - & - & - \\ \hline
\end{tabular}
}
\end{table}

\end{document}